\documentclass[conference]{IEEEtran}
\IEEEoverridecommandlockouts
\usepackage{cite}
\usepackage{amsmath,amssymb,amsfonts}
\usepackage{algorithmic}
\usepackage{graphicx}
\usepackage{textcomp}
\usepackage{xcolor}
\usepackage{hyperref}
\usepackage{siunitx}
\usepackage{balance}
\newcommand\blfootnote[1]{%
  \begingroup
  \renewcommand\thefootnote{}\footnote{#1}%
  \addtocounter{footnote}{-1}%
  \endgroup
}
\def\BibTeX{{\rm B\kern-.05em{\sc i\kern-.025em b}\kern-.08em
    T\kern-.1667em\lower.7ex\hbox{E}\kern-.125emX}}
\begin{document}

\title{How to Do Things with Deep Learning Code
}

\author{
\IEEEauthorblockN{Minh Hua}
\IEEEauthorblockA{Computer Science \\
Johns Hopkins University\\
Baltimore, MD \\
mhua2@jh.edu}
\and
\IEEEauthorblockN{Rita Raley}
\IEEEauthorblockA{English \\
University of California, Santa Barbara\\
Santa Barbara, CA \\
rraley@ucsb.edu
}
}

\maketitle
\thispagestyle{plain}
\pagestyle{plain}

\begin{abstract}
The premise of this article is that a basic understanding of the composition and functioning of large language models is critically urgent. 
To that end, we extract a representational map of OpenAI’s GPT-2 with what we articulate as two classes of deep learning code, that which pertains to the model and that which underwrites applications built around the model. 
We then verify this map through case studies of two popular GPT-2 applications: the text adventure game, \textit{AI Dungeon}, and the language art project, \textit{This Word Does Not Exist}. 
Such an exercise allows us to test the potential of Critical Code Studies when the object of study is deep learning code and to demonstrate the validity of code as an analytical focus for researchers in the subfields of Critical Artificial Intelligence and Critical Machine Learning Studies. 
More broadly, however, our work draws attention to the means by which ordinary users might interact with, and even direct, the behavior of deep learning systems, and by extension works toward demystifying some of the auratic mystery of “AI.” What is
at stake is the possibility of achieving an informed sociotechnical consensus about the responsible applications of large language models, as well as a more expansive sense of their creative capabilities—indeed, understanding how and where engagement occurs allows all of us to become more active participants in the development of machine learning systems.
\blfootnote{First submission, September 2021; revised submission, September 2022; final copy, December 2022 \\
Accepted for publication: \textit{Digital Humanities Quarterly}
}
\end{abstract}


\section{Overview}

\begin{figure}[t!]%
    \centering
    \includegraphics[width=8cm]{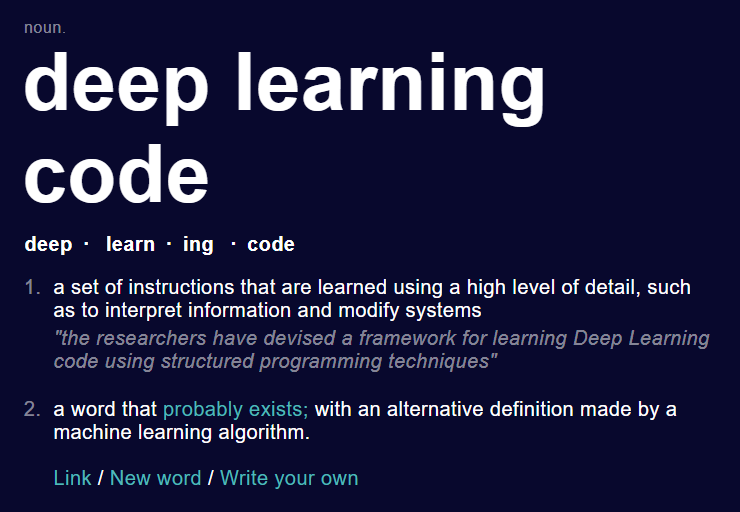}
    \caption{Screen capture from \textit{This Word Does Not Exist}}%
    \label{fig:Picture1}%
\end{figure}

The public conversation about large language models is nothing if not riddled with sensation. 
Even apart from Open-AI’s already-notorious ChatGPT, or a Google engineer’s claim of sentience for its conversational model, the reporting on research advances in Natural Language Processing (NLP) almost inevitably cultivates some degree of alarm and auratic mystery. 
Some of this affect is generalizable to the field of artificial intelligence, which continues to be overcoded as a dangerous and impenetrable black box. But
the enclosure of the NLP research environment behind Application Programming Interfaces (APIs), and the concomitant development of language models requiring compute resources beyond the means of ordinary users, has also contributed to the mystification.\footnote{For example, even though OpenAI’s GPT-3 is now integrated within applications spanning disparate domains and industries, as of this writing it remains accessible to the general public only through an Application Programming Interface (API) \cite{openai2021gpt3nextapps}, and operationally legible only through experiments done by researchers \cite{branwen2020gpt, rong2021extrapolating} or attempts at replicating its source code \cite{wiggers2021aiweekly}.}
Paradigmatic headlines such as \href{https://www.nytimes.com/2022/04/15/magazine/ai-language.html}{“A.I. Is Mastering Language. Should We Trust What It Says?”} communicate the threat of a singular, mystified, and unstoppable thing called “AI,” with humans relegated to the role of end users. 
One salient question for researchers, then, is how best to demystify large language models and explain their functioning, not only to help shape a sociotechnical consensus about their responsible application but also to help expand the horizon of possibility for human engagement, beyond instrumental use and risk assessment alike. 

While there are technical methods to explain and understand language models—so-termed BERTology is one example\footnote{BERTology is a field that studies the specific components of language models such as hidden states and weights. Another example of a project endeavoring to explain language models is Ecco, an open-source library for the creation of interfaces that help to explain language models like GPT-2 by illuminating input saliency and neuron activation \cite{alammar2020interfaces}; a third is \href{https://ml4a.github.io/ml4a/looking_inside_neural_nets/}{Looking inside neural nets}.}—the occasion of a scholarly conversation about Critical Code Studies (CCS) suggests a particular research question: What can be learned about machine learning systems when the techniques and lessons of CCS are applied at the level of code rather than theory? 
As the full range of contributions to this special issue attests, CCS applies critical hermeneutics to software code, documentation, structure, and frameworks. 
Among the more prominent case studies in the aforementioned are readings of BASIC, Perl, JavaScript, and C++ programs, all exercises in excavating the meanings or structures of signification that are both latent in and produced by code [e.g. \cite{montfort201410}]. 
This method has been developed for and honed on \textit{programmed} code: words, characters, and symbols arranged according to rules and in a form that might be understood as textual. 
CCS in other words has traditionally grappled with artifacts that, while certainly mobile and changeable, have enough of a static quality to allow both for individual study and shared understanding. 
What though can this method do with machine learning systems that include code in this ordinary sense as well as statistical parameters and operations, that are in other words less lexical than they are mathematical? 
While we had previously expressed skepticism about the efficacy and utility of taking a CCS approach to a language model such as GPT-2 \cite{hua2020playing}, we were challenged for this special issue to do precisely that: to consider the extent to which CCS might illuminate a system, either in terms of intent or functioning, that is comprised not just of code but also training data, model architecture, and mathematical transformations—to consider then if its methodology might be adapted for the study of an interactive system that is not strictly algorithmic.    

In the wake of CCS, as well as Software Studies and Platform Studies, the subfield of Critical Artificial Intelligence Studies (CAIS) has emerged explicitly to take account of machine learning. 
CAIS calls for an end-to-end engagement and insists on the model as the unit of analysis. 
Even if not self-consciously presented as a field articulation, academic studies of machine learning have collectively shifted the emphasis away from code and toward the model, with a particular emphasis on vectorization, probabilitization, and generalization. 
Adrian Mackenzie, for example, asserts that “code alone cannot fully diagram how machine learners make programs or how they combine knowledge with data” \cite[p.~22]{mackenzie2017machine}. 
And in what may well become a foundational document for CAIS as such, in a so-termed “incursion” into the field, researcher attention is redirected from “an analytical world of the \textit{algorithm} to the world of the \textit{model}, a relatively inert, sequential, and/or recurrent structure of matrices and vectors”  \cite[p.~5]{roberge2021cultural}. 
What this means more plainly is that CAIS concerns itself not with the code that implements a particular machine learning model, but rather its mathematical definition, not with “symbolic logical diagrams” but rather “statistical algorithmic diagrams” \cite[p.~23]{mackenzie2017machine}. 
Deep learning code is thus positioned as merely one component of machine learning systems and not by itself granted priority. 
In contrast, we proceed from the observation that deep learning code in fact represents the myriad possible \textit{implementations} of a machine learning model and its auxiliary programs.
On this basis, we contend that the design choice and biases inherent in each implementation extend their significance beyond mathematical formulae and warrant a closer look.

While the \textit{model} has tended to serve as the unit of analysis, the broader concern of CAIS has been to situate these models in their social, historical, and cultural contexts \cite{mackenzie2015production, burrell2016machine, mackenzie2017machine, underwood2020machine, roberge2021cultural, offert2021latent}.
Linking model architecture to context—whether that be use case, domain of implementation, or institutional setting—has thus allowed CAIS to pose crucial questions about the ethics and politics of machine learning systems. 
So too CCS has endeavored to engage both the socio-historical dimensions and effects of programming code, also with structural transformation as a hoped-for endgame \cite{marino2020critical}. 
What remains to be done, and what is in part the purpose of our essay, is to test both the feasibility as well as the critical potential of CCS as a method when the object of study is deep learning code.

This then is our cue to take a closer look at one of the central terms for this analysis: code. 
While the discipline of Computer Science classifies different algorithms according to their composition and behavior (e.g. binary search, breadth-first search, insertion sort), there is no master “deep learning” algorithm, apart from frequently-used algorithms like backpropagation and domain-specific optimization procedures like gradient descent  \cite{rumelhart1986learning, roberge2021cultural}. 
Deep learning code, then, is not homogenous or singular, a point underscored by the many iterations of single models, e.g. the GPT and DALL-E series. 
In the popular imaginary, deep learning, or AI as a cultural technique, might seem to adhere to the “myth of unitary simplicity and computational purity,” but in actual practice it is a collection of disparate processes that work toward the “production of prediction” \cite{bogost2015cathedral, mackenzie2015production}.

Along these lines, Mackenzie has written of the difficulty of locating the difference between a game console and something like Google’s AlphaGo in program code, suggesting that researchers must look elsewhere to understand the particularities of a machine learning system \cite[p.~22]{mackenzie2017machine}. 
He then goes further to articulate a limit for CCS that we accept as a challenge: “the writing performed by machine learners,” he contests, “cannot be read textually or procedurally as programs might be read.” 
This is the case, he notes, because “the learning or making by learning is far from homogenous, stable, or automatic in practice” \cite[p.~22]{mackenzie2017machine}. 
That CCS as a method is somewhat mismatched or misaligned with technical objects whose behavior is defined by numerical parameters and vectors rather than logical symbols is further evinced by Jenna Burrell’s “audit” of the code of a spam filtering model: as that audit exercise illustrates, seeking to understand the rationale for classification decisions inevitably imposes “a process of human interpretive reasoning on a mathematical process of statistical optimization” \cite{burrell2016machine}.\footnote{For completeness, we remark that machine learning research has traditionally used the concept of inductive bias to describe the set of assumptions that a learning algorithm makes to predict outputs for inputs that it has not previously encountered \cite{alpaydin2020introduction}. For example, the inductive bias of a nearest neighbor algorithm corresponds to the assumption that the class of an instance $x$ will be most similar to the class of other instances that are nearby in Euclidean distance \cite{mitchell2007machine}. Thus, the inductive bias of a learning algorithm loosely corresponds to its interpretation of the training data to make decisions about new data. Using this formulation, one might begin to see how an interpretation of deep learning decision-making might be possible. In contrast, Tom Mitchell observes that “it is difficult to characterize precisely the inductive bias of  BACKPROPAGATION learning, because it depends on the interplay between the gradient descent search and the way in which the weight space spans the space of representable functions. However, one can roughly characterize it as \textit{smooth interpolation between data points}” \cite{mitchell2007machine}. Inductive bias, then, is not the answer to the methodological challenge to which we are responding. The problems of interpretability and explainability are also crucial for deep learning, but not within the immediate purview of this paper. For an overview see \cite{gilpin2018explaining, lipton2018mythos}.}
Deep learning objects that entail parallel processes and training schemes may indeed be challenging to read procedurally as CCS might read JavaScript; however, as we will endeavor to demonstrate, they can nonetheless be read textually and separately from their operations and mathematical foundations.

The common understanding of deep learning code is that it is an implementation of a deep learning model and ancillary functions that support the model, written in a specific programming language. 
Within the scope of our paper, these models are taken to be deep neural networks, specifically the Transformer architecture \cite{vaswani2017attention}. 
In what follows, we will extract a representational map of a particular deep learning model with an open GitHub repository—GPT-2—based on a close reading of two classes of code, that which pertains to the model and that which underwrites applications, thus countering the notion that deep learning code is uniform and homogenous. 
This representational map will in turn draw attention to the means by which we might oversee, interact with, and even direct the behavior of deep learning systems, and by extension both empirically rebut the fantasy of model sentience and demystify some of the auratic mystery of “AI.” 
To test our theory, we present case studies of two popular GPT-2 applications—the text adventure game, \textit{AI Dungeon}, and the conceptual artwork, \textit{This Word Does Not Exist}—both of which demonstrate the interplay between what we will articulate as two classes of deep learning code.

\section{Mapping Deep Learning Code}

\begin{figure*}[t!]%
    \centering
    \includegraphics[width=15cm]{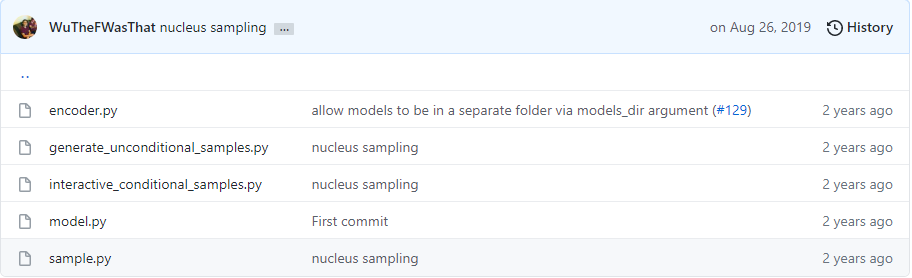}
    \caption{GPT-2’s source files in alphabetical order within its GitHub repository}%
    \label{fig:Picture2}%
\end{figure*}

Our critical study of GPT-2’s source code begins with its GitHub repository and a classification of its code (Figure~\ref{fig:Picture2}). 
Contained in the repository is a README summary, as well as a creative license, requirements file, and a source code folder containing five separate Python files: (1) model.py, as its name suggests, is the kernel of language models such as the GPT series, and it contains functions that define hyperparameters, softmax transformations, attention heads, and a number of familiar deep learning operations; (2) sample.py, also as its name suggests, samples a sequence of text from the model; (3) encoder.py both tokenizes text for model input and decodes text from model output; (4) interactive\_conditional\_samples.py generates samples with an initial prompt; and (5) generate\_unconditional\_samples.py generates samples without an initial prompt.\footnote{To be technically precise, an ‘$\textless\vert \text{endoftext} \vert\textgreater$’ token is used as a way to partition training data and separate different “texts.” Thus, “generate\_unconditional\_samples.py” generates samples from an initial ‘$\textless\vert \text{endoftext} \vert\textgreater$’ token, which is essentially “nothing.”}
Deep learning code, like other software systems, can be metaphorically conceived in terms of layers: the model layer that defines the deep learning model and the application layer that primarily interacts with users or external software. 
Conceptually straddling the model and application layers are sample.py and encoder.py, which perform mediating functions that will become more evident in our subsequent discussion of GPT-2. 
(These latter functions are harder to classify because it remains an open question whether code that samples text from the model or encodes user input and decodes model output is considered part of the model or ancillary functions that support it.) 
What is somewhat surprising—and what in hindsight might have tempered the feverish reaction to ChatGPT in late 2022, had the chatbot been primarily understood as an application or interface—is that there is as yet no common taxonomy or representational map that differentiates the functions and structures representative of deep learning from the functions and structures that are integral, but not particular, to the implementation of deep learning models. 

To that end, we will articulate as a heuristic a practical and conceptual distinction between two classes of code: \textit{core deep learning code} (CDLC), kernel code that defines the deep learning model, and \textit{ancillary deep learning code} (ADLC), ancillary or application code that ordinary developers can replace. 
\textit{Core} deep learning code, of which model.py is the obvious primary instance, has an ordinary and domain-specific meaning: it produces machine learning predictions by implementing deep learning operations.\footnote{For example, matrices feature prominently in core deep learning code because they are used to implement the layers of a neural network. Arithmetic operations are also a frequent occurrence, but are often performed with matrices rather than single numbers. In fact, matrix multiplication is an operation so integral to a neural network’s predictive capabilities that there exists specialized hardware to optimize it.}
Hence, \textit{core} encompasses core data structures and functions that directly execute core deep learning tasks such as classification or regression. 
In contrast, the remaining files in the source code folder operationalize the model’s outputs rather than directly contribute to its predictive functioning. 
They are in this respect \textit{ancillary}, querying or indirectly supporting the model from an external location and, in the case of GPT-2, acting as interfaces that mediate between user requests and deep learning predictions (e.g. specifying and wrangling behaviors, or filtering outputs, as with the aggregating of individual predictions into the composite form of a news article, poem, recipe, or adventure game). 

In this paper, we primarily consider the significance of ancillary code as a post-prediction apparatus, that is, code that is run after a prediction has been made. 
However, it is important to emphasize that ancillary code surrounds the model, and can thus also interact with the model at the start of the production of prediction. 
As we will explain, code such as the web scraper that was used to generate GPT-2’s training data might also be considered as pre-prediction ancillary code with its own politics and significance. 
Although web scrapers can be used for tasks other than compiling training data, those that are used for this purpose are situated within the context of deep learning, and hence can also be considered deep learning code.

Our delineation of the two classes of code is motivated by their distinct empirical and theoretical properties. 
Thus, while deep learning code is often regarded as uniform, even basic—Mackenzie, for example, defines standard machine learning code as “familiar, generic programming” that is “hardly ever hermetically opaque”—our analysis complicates this characterization \cite[p.~26]{mackenzie2017machine}. 
To start, we note that CDLC and ADLC emerge from, and to some degree reflect, different computing practices: artificial intelligence (AI) on the one hand and software development on the other. 
While CDLC is informed by the development of rules for “modifying the connection strengths in simulated networks of artificial neurons” \cite{bengio2021deep}, as well as specialized hardware and software libraries capable of efficiently performing repetitive matrix operations (most notably, AI-accelerated hardware from NVIDIA and Google and software libraries such as Tensorflow and Keras), ADLC in contrast draws from a rich history of web development, APIs, and user interface design, depending on the form that it takes (e.g. a web application that allows users to chat with GPT-2, or an API that allows programmers to send requests to GPT-3 for prediction). 
Given these different historical trajectories, there is enough of a practical difference to explain why CDLC and ADLC seem in their programming conventions, vocabularies, and syntax almost to be different languages, both in their look and how they are read. 
In a nutshell: whereas the complexity of CDLC stems from the complexity of machine learning structures such as neural networks and procedures such as backpropagation, the complexity of ADLC arises from what it does with the machine learning output, such as integrating the outputs into other complex systems—for our purposes, a GPT-2-powered web app that needs to be accessed concurrently by thousands of users. 
A textual comparison of model.py (CDLC) and interactive\_conditional\_samples.py (ADLC) in GPT-2 will help reinforce the distinction. 

\begin{figure}[t!]%
    \centering
    \includegraphics[width=9cm]{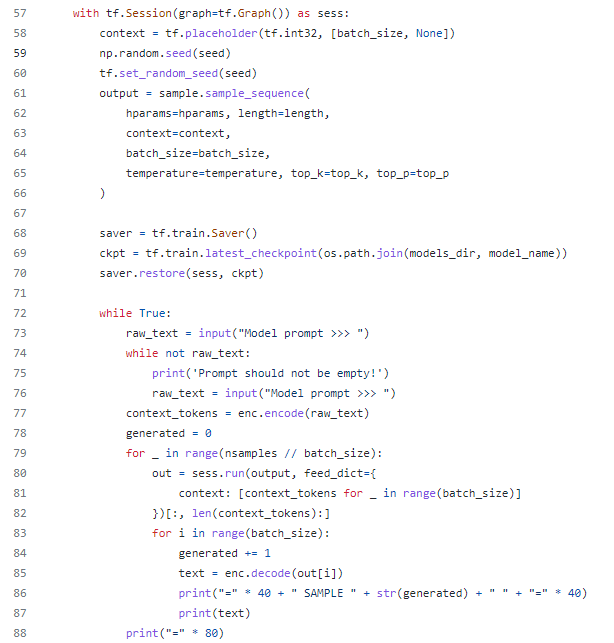}
    \caption{Snippet of interactive\_conditional\_samples.py}%
    \label{fig:Picture3}%
\end{figure}

Whereas interactive\_conditional\_samples.py defines a single function with a specific purpose, model.py in contrast is filled with functions that contain numerous technical and mathematical operations whose meanings may not be initially obvious to non-programmers. 
Although interactive\_conditional\_samples.py starts out with a fair bit of technical setup code, its main loop (lines 72 - 88) clearly communicates the module’s purpose. 
The module prompts for and saves user input into a variable named raw\_text (line 73), encodes (line 77) and feeds it to the model (lines 80 - 82), and decodes (line 85) and prints the output iteratively (lines 86 - 87) (Figure~\ref{fig:Picture3}). 
The translational work ADLC does here is facilitated by the presence of built-in Python functions (input, print) and intuitively-named functions (encode, decode). 
In fact, this syntactic composition is what marks ADLC as such. 
This generic program is easy to understand in GPT-2’s case because it executes basic operations while outsourcing the predictive work to model.py, which handles the complex deep learning work.\footnote{Our claims about the legibility of source code written in Python rely on a distinction between a novice programmer who understands basic principles such as conditional statements and loops and a data scientist who works with neural networks, especially one with a deep understanding ofTransformers and GPT-2.}
We can then, to a certain extent, imagine the execution of interactive\_conditional\_samples.py as we read the code.

In contrast, model.py itself has a complex structure and vocabulary that makes comprehension, and interpretation, especially demanding. 
The imaginative exercise for the execution of model.py’s predictive process is more challenging due to its dependence on vast arrays of numbers and their complex interactions. 
In addition, at any one time the model could conceivably be in several different possible states as determined by its weights, leading to different outputs given a particular input.\footnote{Making a related point, Burrell suggests that it is not productive ultimately to apply interpretation to mathematical optimization, or to attempt to parse the decision-making logic of a neural network—in our terms, CDLC—because of its escalating complexity: “reasoning about, debugging, or improving the algorithm becomes more difficult with more qualities or characteristics provided as inputs, each subtly and imperceptibly shifting the resulting classification” \cite[p.~9]{burrell2016machine}.}
The code looks significantly different from interactive\_conditional\_samples.py because of its almost exclusive dependence on Tensorflow, Google Brain’s machine learning library, which features a host of abstruse-sounding functions, the relative obscurity of which is a necessary product of their specialized and optimized functioning. 
The relative illegibility of GPT-2’s CDLC can partly be attributed to the developers’ programming style and the demanding requirements of production code. 
GPT-2 was written not as a hobbyist project but as a service provider for numerous applications, so while it could have been written using many more loops and basic data structures, it would then have taken much longer to train and run. 
Nevertheless, we can see that the file is arranged into a series of Python functions (softmax, norm, attention\_mask, etc.) and smaller data structures that together form the network representation of GPT-2. 
We can also review the code and derive a sense of the operations being performed (e.g. matrix multiplication, reshapes), but the calculations being performed would be too massive for any one person to parse. 
Even if a reader were familiar with the functions, in other words, GPT-2 is better studied through its mathematical formulation than through its implementation. 
Consider, for example, the line of code “0.5*x*(1+tf.tanh(np.sqrt(2/np.pi)*(x+0.044715*tf.pow(x, 3))))” to describe GELU (an activating function frequently used in neural networks), which is much easier to parse if read as a math formula.

The complexity of model.py, and more broadly the inability to fully examine its execution, does not however mean that one cannot study the intricacies of deep learning systems, either technically or critically. 
For this purpose, we can take a cue from computer science and apply the concept of abstraction to our analysis. 
Because fundamental deep learning operations such as vectorization, learning, and gradient descent exist in the mathematical abstract without programming implementation, it is possible to talk about them without citing the underlying code.\footnote{See \cite{berner2021modern} and \cite{roberts2022principles} for an emerging mathematical theory and theoretical treatment (respectively) of deep learning. See \cite{alpaydin2020introduction} and \cite{mitchell2007machine} for core texts on machine learning and deep learning.} 
One could make the argument that all computer code subtends abstract math, but the challenge of understanding CDLC has meant that abstraction has almost necessarily been at the core of the study of deep learning operations. 
For example, research on machine learning models has thus far focused on the diagram \cite{mackenzie2017machine}, vectorization \cite{mackenzie2015production, parrish2018vectors}, learning \cite{roberge2021cultural}, and pattern recognition \cite{mackenzie2015production, roberge2021cultural}. 
In contrast, the abstract counterpart of ADLC is not prominently featured in analyses of deep learning. 
Put another way, the possibility of analyzing model.py and its operations analogically and metaphorically has already been proven by extensive qualitative engagement with deep learning. 
While we by no means wish to dismiss or even contest the project of CAIS—abstracting deep learning operations is after all foundational to the work of interpretation—we can nonetheless ask what is missed if researchers only focus on training data, model architecture, and the abstraction of deep learning code. \footnote{Training data is one of the more urgent, as well as more visible, problems necessitating critical intervention, of which Joy Buolamwini and Timnit Gebru’s work is particularly of note \cite{buolamwini2018gender}.}

Although we have suggested that ADLC looks like generic programming, it is important to note that its substance and significance differs by operational context. 
As a thought experiment, consider the loop, the signature feature of which, as Wilfried Hou Je Bek explains, “is the minimal means that result in automated infinite production” \cite{bek2008software}. 
This infinite production is trivial in the context of a small computer program that prints out numbers from one to ten, and has historical precedents in automation and mechanical reproduction, but takes on a more sinister value in the contexts of disinformation, hate speech, or even autonomous weaponry. 
These then are the stakes for a CCS analysis of ancillary deep learning code, which operates outside the periphery of a deep learning model but is fundamental to its operation. 

Our premise, then, is that analysis grounded in code citation has perhaps too quickly been deemed inessential to the work of understanding and engaging artificial intelligence. 
More specifically, our suggestion is that ADLC is productive terrain for critical and creative engagement, not least because its structure, logic, and syntactic form is closer to human language than CDLC.\footnote{There may well be examples of creative developers who have tried to bend the rules at the level of the model in the creation of innovative deep learning structures and operations, e.g. the attention mechanism in the Transformer network. It is also perfectly reasonable to expect to see interesting code comments in both CDLC and ADLC (although in our experience it is not common for the former). We stand by our initial observation that it would be much more difficult and time-intensive to apply the CCS framework to study CDLC, but we also think it would be a generative exercise for future research. For this paper, however, we choose to focus on what is in our view the more manifestly creative area of deep learning code: ADLC.}
While the predictive outputs of CDLC, particularly model.py, are in the form of unprocessed numerical data, ADLC does the heavy lifting of translating and assigning meaning to that numerical data.\footnote{Consider the example of a classifier like Logistic Regression trained to classify tumors: depending on the data, the two outputs might be 0 or 1, which would not be meaningful without assigning them respective values of benign and malignant. Such a classification would be necessarily arbitrary and this serves as a reminder of the real interpretive work that happens in the use of ADLC to transform numerical output into linguistic form. This notion of transformation is especially compelling in the context of deep learning because so many transformations have to take place in order for a prediction to be made. Data extracted from an external source must be cleaned and formatted before it can be used to train a model, and an input requiring a predicted output must also be cleaned and would be transformed into numerous vector spaces as it is fed through the model.}
Because CDLC replicates or maps mathematical formulas and deep learning concepts (e.g. code that implements a layer of a neural network), the interpretive work is seemingly limited to the implementation details, as in the choice of framework or programming language. 
ADLC, on the other hand, can be used to interpret and manipulate the raw output in seemingly unlimited ways, and it is this creative and ethicopolitical potential that makes it a particularly fertile area for CCS analysis. 
In the following sections, we trace examples of ADLC doing this transformative work with increasing levels of complexity and identify two of its major functions in the context of a text adventure game and a fictional dictionary: structuring the output from GPT-2 and constructing interfaces that mediate between the user and the language model. 

Our analysis starts with a closer look at GPT-2’s ADLC to help illustrate the argument and suggest possibilities of thinking both logically and intuitively about it. 
ADLC is an accessible, and replaceable, component of GPT-2’s source. 
Removing the two files, generate\_unconditional\_samples.py and interactive\_conditional\_samples.py, would thus not affect GPT-2’s production of prediction, and in fact most forks of GPT-2’s source code replace the files with the programmer’s own interface. 
If considered in relation to word vectors and the attention mechanism, generate\_unconditional\_samples.py’s printing of GPT-2’s outputs might be regarded as a somewhat fundamental and trivial operation. 
But printing GPT-2’s output is only one implementation of ADLC’s mediation between user prompts and parameters (length of sample, number of samples, randomness) and model parameters (length, nsamples, temperature)—and other implementations, other forms of interaction, as we will outline, are both more interesting and more generative. 
Regarding GPT-2 as an interactive system rather than an algorithm, with ADLC as the front-facing interface for an underlying language model, allows us to explore the possibilities as well as limits of ADLC with respect to controlling GPT-2’s behaviors. 
That is, the behavior of GPT-2, as the user experiences it, can be understood to be configured by its myriad interfaces rather than its internal predictive logic engine, which is itself only limited to one behavior: text generation. 
These interfaces extend GPT-2’s behavioral capacity and transform text generation into media generation, interactive entertainment, and collaborative writing. 
Put another way, although GPT-2 is a language model that generates text sequences, it takes specific examples of ADLC in order for GPT-2 to function as a news writer, dungeon master, chat bot, or language artist, as it is imagined to do for \textit{AI Dungeon}, \textit{This Word Does Not Exist}, and many other applications besides.

It is perhaps an understatement to note that the playful quality of these two applications is somewhat at odds with the instrumentalist quality of GPT-2’s ADLC. 
For a field such as CCS, which has historically been attuned to the rhetorical properties of code and to what makes it interesting as a cultural artifact—not only its literariness and humor but also its political potential, one excellent example of which is the CCS working group’s discussion of gender and \href{http://wg20.criticalcodestudies.com/index.php?p=/discussion/18/week-1-colossus-and-luminary-the-apollo-11-guidance-computer-agc-code}{The Apollo 11 Guidance Computer (AGC) Code}—the relative dryness of GPT-2’s source might initially cause some perplexity. 
There are no unreadable texts, to be sure, but there might seem to be limits to what can be said about numerical data and code that bears no traces of programmer whimsy and is absent not only of cultural references, in-jokes, and easter eggs, but also of manifest traces of authorship and subjectivity, apart from the occasional typographical error. 
Counter-intuitive as it may seem, however, even a snippet of ADLC can be remarkably generative. 
If, for example, the code for a mobile phone application can be read as an instantiation of an argument about immigration and border politics \cite{marino2020critical}, we might by extension consider how a line such as “if 0, continues to generate samples [indefinitely]” opens up philosophical and political questions of control.   

\begin{figure}[t!]%
    \centering
    \includegraphics[width=9cm]{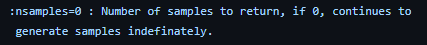}
    \caption{Snippet in the sample\_model function within generate\_unconditional\_samples.py}%
    \label{fig:Picture4}%
\end{figure}

\begin{figure*}[t!]%
    \centering
    \includegraphics[width=18cm]{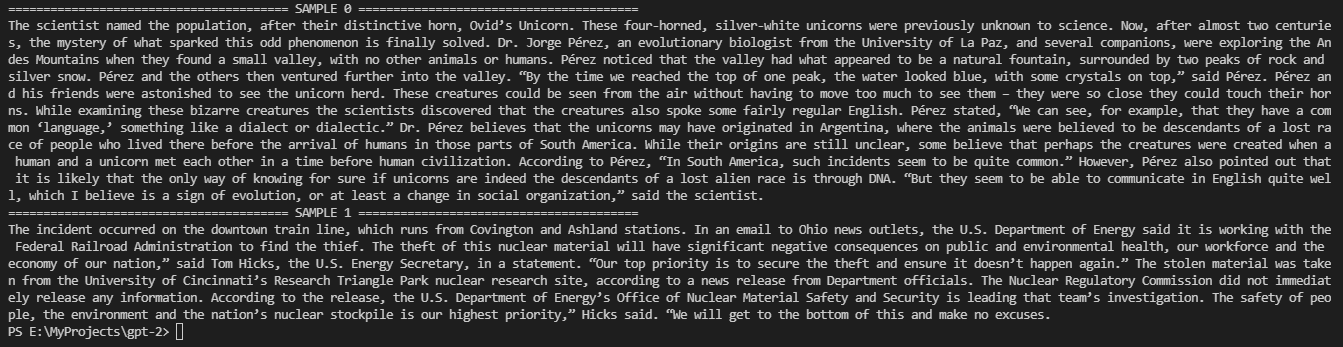}
    \caption{Reconstruction of GPT-2’s outputs printed to the console}%
    \label{fig:Picture5}%
\end{figure*}

This snippet of generate\_unconditional\_samples.py comes from the function sample\_model’s docstring and explains how the nsamples parameter is used to specify the number of samples that GPT-2 generates (Figure~\ref{fig:Picture4}).\footnote{GPT-2’s internal model of the English language relies on the assumption that the current words in a sentence determine the occurrence of subsequent words \cite{bengio2008neural}. At each prediction time step, GPT-2 outputs the probability of every textual unit in its vocabulary (50,257 for the smallest model) given the previous words in a sequence. While a common observation of GPT-2 in action might be that it is “writing,” then, what it is actually doing is “sampling,” by which we mean the drawing of a random observation from a probability distribution. }
Dictated by Python conventions, a docstring is a block of text nested within a function that acts as an explanatory summary, and thus can be studied under the same critical apparatus as software paratexts and code comments \cite{goodgerrossum2001docstring, douglass2010comments}. 
The snippet is taken from the file, generate\_unconditional\_samples.py, which allows for the most rudimentary form of interaction between a user and GPT-2. 
After configuring a few parameters such as random seed, model size, and length and number of samples to generate, and then executing the program, generate\_unconditional\_samples.py will continue to print samples to the console in the format below until the maximum number of samples specified is reached (Figure~\ref{fig:Picture5}). 

\begin{figure}[t!]%
    \centering
    \includegraphics[width=9cm]{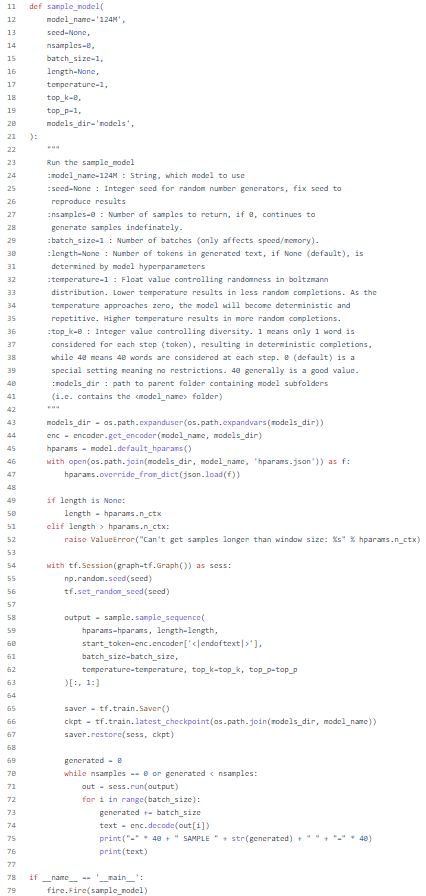}
    \caption{Snippet of generate\_unconditional\_samples.py}%
    \label{fig:Picture6}%
\end{figure}

This prescribed limit is particularly suggestive because the original developers effectively designed a way for it to be circumvented. 
As line 70 of the code indicates, GPT-2 will continue to predict and print samples unless one of two conditions is true: (1) the number of samples generated exceeds the “nsamples” limit or (2) the “nsamples” limit supplied is zero (Figure~\ref{fig:Picture6}). 
Thus, if “nsamples” were zero, GPT-2 would print samples indefinitely until interrupted by the user’s keyboard, memory depletion, or power loss.\footnote{\href{https://docs.python.org/3/library/stdtypes.html\#numeric-types-int-float-long-complex}{Python} can handle integers with effectively “unlimited precision,” which means that there is not a limit on how big the value for “generated” can be, unlike languages such as Java and C \cite{golubinpython}. What this means is that “generated” can exceed the conventional limits on integers, so that the main loop in sample\_model can indeed run indefinitely.}
(The parallels with Stanislaw Lem and Trurl’s machines write themselves.)

Nevertheless, the irony of zero as a limit for infinite generation has the effect of drawing our attention to the crux of the matter: the point at which human action affects—or fails to affect—machine behavior. 
In this particular instance, the interaction with GPT-2 is not only limited but essentially one sided; in technically imprecise terms, it amounts to ‘start the machine and let it run’. 
It might even be conceived as a contemporary iteration of the “Hello, World!” program, with the text passed onto the print function generated by another function (GPT-2) rather than a human. 
It is perhaps an obvious point but one that bears repeating: interactions between humans and machine learning systems cannot and should not take this form.

If restricted human action is suggested by the syntax of the file name, ‘generate unconditional’, then ‘interactive unconditional’ might seem in turn to promise a more substantive mode of engagement than mere behavior initiation. 
Indeed, interactive\_unconditional\_samples.py does make it possible for users to “interact” and have a degree of influence over GPT-2’s predictions, in that several lines of this code allow users to input a phrase or sentence of their choosing for the model to complete. 
But here too action is limited, in this instance to the production of single-use seed text: thus, type “the quick brown fox” and then wait for the output. 
GPT-2 will attempt to follow from the prompt and complete the sentence as many times as specified, but each iteration is distinct in that the user needs to prompt the model with more seed text and the model will not take prior input into account when generating further text. 
While the call-and-response structure of an interaction, for example, might seem to suggest a balance of forces, and even a kind of mutuality between human and machine, the process is better thought of in terms of input-output, with all its connotations of routinized, mechanized, even roboticized action.\footnote{The high quality of ChatGPT’s output, and particularly its ability to remember prior exchanges, seems once again to have invoked the specter of machinic liveness, but the structure of the input-output process with GPT-3.5 is of course no different from GPT-2.}
It is in other words cooperative writing only in the rudimentary sense that there are at least two actants generating words in sequence.

\begin{figure}[t!]%
    \centering
    \includegraphics[width=6cm]{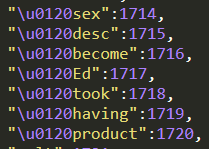}
    \caption{Representative entries in the dictionary used for encoding and decoding. After the special unicode token, “$\backslash$u0120,” the words are followed by their numerical ID.}%
    \label{fig:Picture7}%
\end{figure}

We can now shift our attention to the degree with which generate\_unconditional\_samples.py and interactive\_conditional\_samples.py transform the language model’s numerical output and remark on their equally basic operation. 
(For economy, we restrict our attention to generate\_unconditional\_samples.py, as the analysis applies to both.) 
After querying the model and receiving the results, generate\_unconditional\_samples saves the output to a variable succinctly named “out” on line 71. 
At this point, the data contained in “out” would be meaningless to readers because it is still numerical; that is, each individual token in the output string is still encoded as numbers. 
In order to transform the data to readable text, it must then be passed to a decoder on line 74, at which point the decoding mechanism housed within encoder.py looks up a dictionary entry to map the numerical tokens back to their textual counterparts (Figure~\ref{fig:Picture7}). 
Although the translational work here is non-trivial—readers cannot otherwise parse GPT-2’s output—the operation itself is basic and limited to dictionary queries. 
It is though of course possible to use ADLC to perform more complex transformations after the decoding stage to more dramatically alter GPT-2’s output, as we shall outline. 

This deep dive into GPT-2’s GitHub repository and analysis of its ADLC has thus far not specified the \textit{which} and the \textit{when}—all-important delimiters for a language model that came to fame in part because of its staged release \cite{solaiman2019release}. 
We have hinted at the reconstruction required to look back and analyze the experience of what it was like to interact with GPT-2 in 2019 and can now be more explicit about the archaeological work one has to do in order to analyze the model in the present. 
What we have studied thus far is an almost-exact snapshot of the ADLC as it appeared in the first release in 2019, a picture that represents OpenAI’s initial vision and its open invitation to developers to explore the limits of GPT-2’s behavior. 
The intervening years have seen a number of changes, not only to GPT-2 but also to its dependencies. 
Tensorflow, for example, has evolved to such an extent that much of the code of GPT-2 will not work unless updated to align with Tensorflow’s new protocols. 
Even more important to the running of the original GPT-2 model are the weights, which must be downloaded via “download\_model.py” in order for the model to execute. 
Thus, one could envision a future CCS scholar, or even ordinary user, attempting to read GPT-2 without the weights and finding the analytical exercise to be radically limited, akin to excavating a machine without an energy source from the graveyard of “\href{http://www.alamut.com/subj/artiface/deadMedia/dM_Manifesto.html}{dead media}.”

With an analysis of interactive\_conditional\_samples.py, we seem to have come to a fork in the road of developer engagements with GPT-2’s source code. 
We have mapped its structure and identified the prescribed interfaces through which the model interacts with the world, observing the relatively basic means by which GPT-2’s ADLC transforms the raw output and constructs the interaction medium. 
The next step takes us outside the repository so we may consider how the circulation of GPT-2’s source code generated new instances of ADLC that in turn morphed into new interfaces for interaction and new forms of management, regulation, and control. 

\section{Applications of GPT-2}

GPT-2 can be described in basic terms as a computational procedure that takes a vectorized token (a textual unit) as input and outputs a corresponding vector of token probabilities. 
Extensive user engagement, however, has transformed the language model into an interactive object whose function extends well beyond the production of deep learning prediction.\footnote{For this issue of interaction, we find helpful Florian Cramer’s articulation of software as “a cultural practice made up of (a) algorithms, (b) possibly, but not necessarily in conjunction with imaginary or actual machines, (c) human interaction in a broad sense of any cultural appropriation and use, and (d) speculative imagination” \cite[p.~124]{cramer2005words}. We also find helpful the genealogical account of computer interaction in \cite{murtaugh2008software} as well as the account of a paradigm shift from algorithms to interactive systems in \cite{wegner1996paradigm}.} 
The story of the staged release of GPT-2 is at this point perhaps well known: it proceeds from the debut of the partial model (117m parameters) in February 2019, its fanfare fueled both by the fantastic tale of Ovid’s unicorn as well as the suggestion that it performed too well and thus needed to be guarded against malicious use through the withholding of the full 1.5b parameter model \cite{solaiman2019release}. 
Because the release was open source and the technological barrier for interaction relatively low, the announcement of GPT-2 as a “better model” was a siren song that lured users of all capacities into a range of experiments to test its capacities and the code went into widespread circulation. 
The release then was a textbook instance of a computational object moving from the “clean room of the algorithm” to the “wild, tainted and unpredictable space of dynamic and shared environment” \cite[p.~148]{murtaugh2008software}. 
Marino narrates such a trajectory in related terms, writing of his realization that he needed to “take code out of the black box and explore the way its complex and unique sign systems make meaning as it circulates through varied contexts” \cite[p.~19]{marino2020critical}. 
In the case of GPT-2, much of the enthusiasm driving circulation derived from experiments with fine-tuning, the process of editing GPT-2’s internal weights by training it on specific texts so that it can better emulate them. 
Fine-tuning alters the CDLC and might be thought as a form of technical speculation (e.g. altering the parameters and the training data). 
Although we will not elaborate on how fine-tuning affects the model, as we shall see shortly, the process plays an important role in enabling \textit{This Word Does Not Exist} to generate fictitious dictionary entries. 
Our concern will be a more expressive and conceptual form of speculation, in the guise of transforming ADLC and developing different interfaces that allow for a modicum of control over GPT-2. 

Perhaps the most organic extensions of the form of interaction made possible by interactive\_conditional\_samples.py are projects that configure GPT-2 as a writing assistant. 
Interactive writing platforms such as “\href{https://transformer.huggingface.co/}{Write With Transformer}” situate GPT-2 in an auxiliary role as a kind of editor, queueing up the model to autocomplete the user’s input text, and thus reaffirm a hierarchy of labor (it is after all the user who ‘writes with’ GPT-2). 
TabNine, which hooks into conventional code editors like Visual Studios Code, is particularly notable in this regard in that GPT-2 is pulled into an authoring environment in which the user can direct its operations, as opposed to writing with GPT-2 within its source code. 
While certainly functional, these examples are precisely and even only that: although there is a certain degree of fine-tuning that takes place to “shape” GPT-2’s output, the language model’s output is precisely replicated for the user without any addition modifications that might otherwise improve its quality (which we have already seen with generate\_unconditional\_samples.py and interactive\_conditional\_samples.py). 
For more imaginative work with ADLC that treats GPT-2’s output not as plain text but as poetic material, and interaction not as a basic chatbot session but as something more like an expansive \textit{Dungeons \& Dragons} session, we turn to \textit{AI Dungeon} and \textit{This Word Does Not Exist}, two of many implementations that demonstrate the transformative work of developers outside of OpenAI and attest to the vibrancy and collaborative spirit of the NLP community writ large. 

Originating in a college hackathon project, \textit{AI Dungeon} is built on a version of the GPT-2 model fine-tuned on a corpus of choose-your-own-adventure stories. 
While we have elsewhere elaborated on the game’s provision of a framework for evaluating language models and its status as a paradigmatic instance of “citizen NLP,” for this analysis we wish to highlight the political and aesthetic potential of the commands available on the game’s interface as they are articulated by ADLC \cite{hua2020playing}. 
Particularly when viewed in relation to the modes of interaction we have here been charting, game commands like “remember,” “revert,” and “alter” serve as potent examples of true collaboration between human users and deep learning models. 
To start, we return to the GitHub repository for an earlier version of \textit{AI Dungeon}, in order to highlight the ADLC responsible for the game interface \cite{walton2019aidungeon}. 
Within the repository, there is a copy of the GPT-2 model in the generator folder, which also contains CTRL, another language model with which the game was experimenting. 
The GPT-2 in this generator folder is almost a mirror copy of the one in the original GPT-2 repository, apart for the notable absence of the aforementioned ADLC modules—generate\_unconditional and interactive\_conditional—which have here been replaced by play.py, a more expressive form of ADLC structured as a text adventure game that feeds user inputs to a language model for the response. 

play.py contains code that prints out a “splash image” of the game’s title and sets up the archetypical “$>$” delimiter awaiting user input. 
The module initiates an iterative gameplay loop whereby the underlying language model narrates a story, awaits for user input in the linguistic form of actions or game commands, processes the input, and then returns the next portion of the story. 
The call-and-response loop with GPT-2 is familiar, but with subtle differences: play.py calls GPT-2 via two wrappers that extend the language model’s functionality on lines 175 and 176. 
Called on line 175, GPT2Generator essentially mimics generate\_unconditional\_samples.py or interactive\_conditional\_samples.py by processing the user input and querying the underlying language model. 
However, GPT2Generator extends the language model’s functionality by doing some post-processing work such as removing trailing sentences, converting the output into the second person, correcting punctuation (at line 85 in gpt2\_generator.py, it moves periods outside quotation marks), and filtering so-termed bad words. 
What GPT2Generator does then is somewhat inexactly shape the model’s output to bring it more in line with contemporary linguistic protocols and the generic conventions of a text adventure game. 
The second wrapper, StoryManager, is built on top of GPT2Generator and does the work that its name suggests: past events generated by GPT-2 and the player’s actions are strung together and maintained in memory, constituting a “story.”

While work that play.py does to transform or translate the raw output from GPT-2 into the form of a text adventure game is fundamental, we find even more compelling the use of ADLC to extend the underlying language model’s functionality and radically reimagine our interaction with it. 
Of particular note are those mechanics that, true to the form of a game, prompt incremental action and in so doing emphasize collaboration and revision. 
“Remember,” for example, is an experimental command allowing players to embed pieces of information that are constantly fed into GPT-2 at each step of the prediction and thus emulate a loose “memory” for the AI narrator.
“Revert” functions as a kind of rewind and allows players to return to any step in the preceding narrative. 
“Alter,” perhaps the most dramatic of the three, allows for a complete edit of the AI’s predictions at each step, offering players the remarkable ability to address and fix predictions, as opposed to prompting the model to produce a new prediction from scratch, as if no lessons had been learned. 
While these commands all give some indication of the expressive capacity of next-gen text generators and advance the development of different modes of human-AI interaction, what is especially germane to our discussion are the processes and conditions that made them possible in the first place. 

Notably, such creative work is enabled by very basic programming constructs. 
Conditional statements on lines 233 to line 320 in play.py determine what command the user wants to take (e.g. ‘elif command == “revert”’). 
Within these conditional statements, rudimentary operations are creatively exploited to mimic advanced behaviors; for example, reverting a game step is as simple as decrementing an index within the story manager’s memory. 
Other complex mechanics like “alter” might be enabled by more complex functions, but what we highlight here is the relatively low bar for engagement with language models and deep learning application development. 
In the case of AI Dungeon, almost all of the extra behaviors for the game stem from commands coded in play.py, an instance of ADLC in the “wild” \cite{murtaugh2008software} and the means by which developers of all abilities, including those who are not specialists in machine learning, can tinker with GPT-2, manipulate its outputs, and more broadly operationalize the model so that it serves as a medium for creative production.\footnote{ Python creator Guido van Rossum states in an interview that the programming language was implemented in a way that emphasizes interactivity and that it was based on ABC, which was “intended to be a programming language that could be taught to intelligent computer users who were not computer programmers or software developers in any sense” \cite{venners2003makingofpython}. Deep learning’s commitment to Python thus had the effect of expanding access to a wide population of programmers. Coding in such an easily-accessible and common language (overshadowed only by JavaScript and HTML/CCS) and opening the source code (model.py) to be callable implicitly invites other Python users to interact and experiment with your model.}
Improving GPT-2, or any deep learning model, certainly depends on the modification of model architecture and the development of further training data, but the work done with play.py suggests that ADLC is a no less meaningful means by which to direct the model’s functioning—all the more so because it is a technique and tool available to ordinary users.

\begin{figure}[t!]%
    \centering
    \includegraphics[width=7cm]{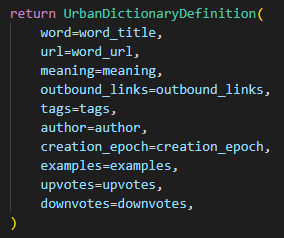}
    \caption{The data structure that constitutes a dictionary entry}%
    \label{fig:Picture8}%
\end{figure}

Borrowing its title from the notorious deep fake application, \textit{This Person Does Not Exist}, \textit{This Word Does Not Exist} (TWDNE) serves as a further example of the creative capacities of ADLC, as well as, not coincidentally, the discursive framing of GPT-2 as “dangerous” \cite{smith2020dangerous}. 
Generating a faux lexicography requires two steps: conditioning (fine-tuning) the model and post-processing the output. 
Conditioning starts with urban\_dictionary\_scraper.py, which, as its name suggests, scrapes Urban Dictionary and formats the information into a data structure that corresponds to a dictionary entry \cite{dimson2020thisworddoesnotexist}. 
The function \_parse\_definition\_div then scrapes the site using the common method of searching for HTML elements corresponding to the specified information—in this instance, finding the location of the word’s definition and then looking for HTML elements with an attribute appropriately named “meaning.” 
Once all the relevant information is extracted, it is inserted into a data structure that organizes the information for final compilation into a fine-tuning dataset (Figure~\ref{fig:Picture8}). 

After the relevant dictionary entries are scraped, they are collated and formatted in train.py into a training dataset used to fine-tune GPT-2. 
Of note in this process is the clever communication protocol that facilitates information extraction down the line. 
On line 903 in train.py, special tokens ($\textless\vert \text{bod} \vert\textgreater$, $\textless\vert \text{pos} \vert\textgreater$, $\textless\vert \text{bd} \vert\textgreater$, $\textless\vert \text{be} \vert\textgreater$, $\textless\vert \text{pad} \vert\textgreater$) are inserted and divide the training instances (i.e. a word and relevant information like its definition) into informational chunks. 
In this way, GPT-2 is conditioned to generate strings of text that have the embedded structure of a dictionary entry, with a word, part of speech, definition, and exemplary usage. 

\begin{figure}[t!]%
    \centering
    \includegraphics[width=9cm]{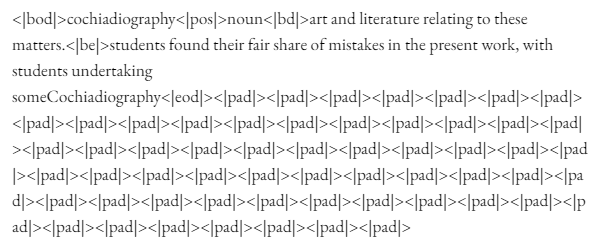}
    \caption{Raw output of the word "cochiadiography"}%
    \label{fig:Picture11}%
\end{figure}

The embedded structure is evident in the example of the custom generation of the fictitious word, “cochiadiography,” which begins with word\_generator.py to gather user input and initialize parameters, and then datasets.py to call on GPT-2 for text generation on line 567 (GPT-2 is here also enclosed in a wrapper class that extends its functionality) (Figure~\ref{fig:Picture11}). 

In Figure~\ref{fig:Picture11}, we see that the relevant information is cleverly separated by artificial delimiters designed by the programmer. 
After the raw output is generated, a pattern matching process starting on line 586 of datasets.py further decodes the output. 
(Note that this human-readable string implies that a decoding step that was highlighted in generate\_unconditional\_samples.py has already taken place.) 
On this line, the output is matched against a predefined pattern to identify the aforementioned special tokens that articulate the information in a dictionary entry, as well as information nestled between the tokens (with this specification, the ‘$\textless\vert \text{pad} \vert\textgreater$’ token will be regarded as meaningless and discarded). 
Next, another pattern matching function (“group”) parses the successful string to extract the different pieces of information and save them to separate variables. 
At this point, other post-processing measures akin to those used by \textit{AI Dungeon} are in play, notably among them the filters that catch and remove objectionable content. 

\begin{figure*}[t!]%
    \centering
    \includegraphics[width=15cm]{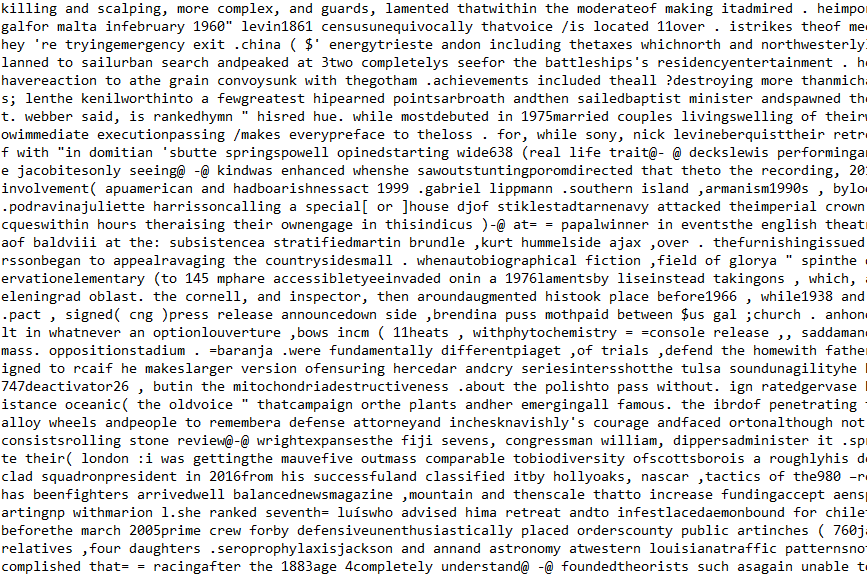}
    \caption{TWDNE’s blacklist}%
    \label{fig:Picture9}%
\end{figure*}

Removing profanity from \textit{AI Dungeon} is a basic operation by no means particular to deep learning: unless “censorship” is toggled off, output from GPT-2 is passed to the aptly named remove\_profanity function, which relies on an external module, also aptly named ProfanityFilter, to replace words appearing on a custom censor list with strings of the ‘*’ character. 
That there is something inherently subjective and even arbitrary about a list of 114 profane words should be evident from the start but further evinced by the absence of “goddamned”; that there is something blunt about the operation is evinced by the use of the Python “re” (regular expression) for basic find-and-replace. 
More suggestive because more intricate is TWDNE’s screening for hate speech with a custom blacklist of some 5,763,399 words and phrases that begins as follows when a set of blacklisted words is loaded (Figure~\ref{fig:Picture9}). 

What garbage conceptual poetry is this, a reader might ask, and they would not be wrong to wonder at the sourcing and means by which such a text file would be generated. 
The atrocities grossly alluded to at the outset (“killing and scalping”) are clear enough, and reference to Levin Tilmon writing to Abraham Lincoln in 1861 about the composition of the Union forces might lead one to surmise that the filter flags all content related to the U.S. Civil War, but what exactly happened in Malta in February 1960, and would this determination necessitate a deep dive into the fevered swamps of 4chan’s /pol/ board, or at least a passing familiarity with contemporary conspiracy theory?\footnote{TWDNE’s blacklist is used not only to filter output but also to evaluate the creativity of the same:\\
“creative words”: 1 – (num\_blacklisted / max(num\_succeeded\_match, 1)), \\
“nonconforming”: num\_failed\_match / num\_generated}
And what to do with “more complex,” also a delimited token, much less “and guards”? 
What taboo subjects or language do these expressions index and why should they be subject to content moderation? 
Both the content and the form of the .txt file, with all of the liberties taken therein with punctuation and spelling, suggest that the list results from either model training or finetuning, with “and guards” perhaps appearing, who knows, in an Urban Dictionary entry that also mentions the Holocaust (and indeed the paradigmatic logic of machine learning—correlationism—is here very much on display). 
To use the Saussurean framework, we might note then that ADLC not only resituates numerical data within the realm of \textit{langue} (a syntactical, rules-based conception of language), as we have seen with the decoding mechanism in generate\_unconditional\_samples.py, but in the case of “bad word” filters, it also manifestly pulls that output into the realm of \textit{parole} (gesturing toward, if not always adhering to, social conventions in its treating of certain words and concepts as taboo).     

\begin{figure}[t!]%
    \centering
    \includegraphics[width=8cm]{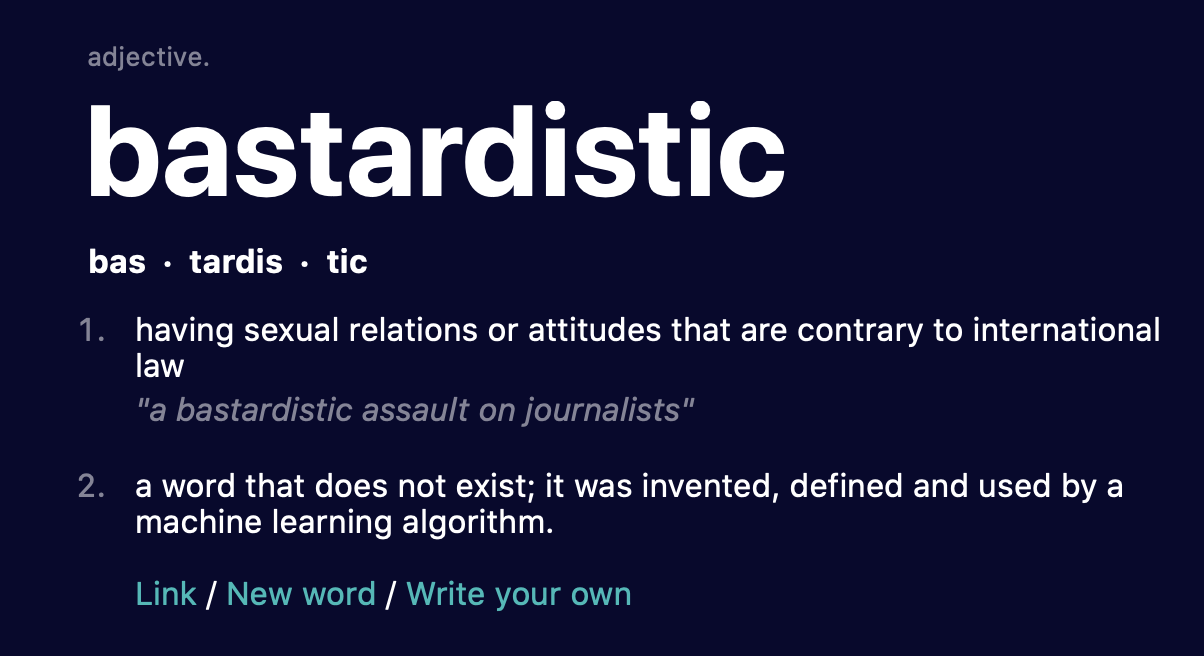}
    \caption{Screen capture from TWDNE}%
    \label{fig:Picture10}%
\end{figure}

Regardless of the precise mode of composition of the blacklist used by the faux lexicographic application, its arbitrariness is evinced by our discovery of the word, “bastardistic” (Figure~\ref{fig:Picture10}). 
Like “cochiadiography,” “bastardistic” began as a string generated by the language model. 
To borrow terms from TWDNE’s template, that string was then “invented, defined and used” as a word—more properly, translated into a word—by datasets.py and not filtered by the blacklist, a process that opens up another meaning of the project’s title. 
It is not simply that “bastardistic” is a representative word that does not exist, but that the application is not actually outputting words; what is at work, in other words, is not language generation, but rather function returns. 
Good output, by whatever metric, might summon the fantasy of model sentience, but this is easily countered by an interface such as TWDNE, which engineers the output in a manner that exposes GPT-2 as a mere function returning four pieces of structured lexicographic data. 

Part of OpenAI’s stated rationale for the withholding of the full model of GPT-2 in February 2019, we recall, was concern about prospective misuse, which would include not only disinformation but also hate speech \cite{radford2019better}. 
There was then and is now work being done at the level of training data and the model (CLDC) to guard against the emergence of another Microsoft Tay, but it is nonetheless left to developers to determine the acceptable parameters of their applications and implementations of GPT-2. 
Pattern matching, regular expressions, and filtering—all examples of ADLC performing ordinary functions in relation to deep learning objects—are some of the means toward this end. 
The need for such ad hoc techniques at all stages of the communication pipeline, however, serves as a reminder that the wild expressivity of NLG can take both negative and affirmative form—hence the discourse on risk and responsibility. 
Whereas generate\_unconditional\_samples.py in the GPT-2 source code functions with something like complete openness, then, it falls to the ADLC in \textit{AI Dungeon} and \textit{This Word Does Not Exist} to manage and regulate the model’s output. 

\section{Conclusion}
Through the developmental arc extending from generate\_unconditional\_samples.py to AI Dungeon and This Word Does Not Exist, we have traced a process of continual modification and refinement of GPT-2’s behavior through distributed engagement, the effect of which has been to bring the model into true cooperative relation. 
Our analysis of GPT-2’s code has shown that deep learning’s potential for interactivity is embedded within the open-source Python kernel that made widely available the tools for building interfaces that structure user engagement. 
The open source code also made it possible for developers to identify and then address unacceptable or otherwise faulty behavior—exercises that arguably contribute to a sense of community and collective responsibility. 
While the source code defining the limits of interaction was written by OpenAI, then, the whole of the code apparatus that we now refer to as “GPT-2” was authored by a wide range of developers, from amateurs to academics and professional programmers.\footnote{EuletherAI’s open-source GPT-J is particularly noteworthy here, not only as a model for responsible AI research but also as an impressive ‘outsider’ appropriation of the model.}
At some level, the centralization of AI research cannot be contested, not least because of the massive quantities of data available to, because extracted by, companies such as Google, Facebook, Amazon, and OpenAI. 
But from a different angle the research environment seems far more distributed, with (often) self-trained individuals and institutional actors alike working to refine the behavior of deep learning models. 

Future CCS work with language models can continue work with GPT-2 as well as other open-source models such as the permutations of GPT and BERT, BLOOM, CLIP, and DALL-E.\footnote{While our analysis holds for GPT-2’s ADLC, further research is required in order to determine whether it applies to other domains such as computer vision and generative adversarial networks (GANs).}
It can also lend its voice to the conversation about what can and cannot be done with GPT-3’s API and to the request to make the model more easily callable by either external software or non-software products. 
If the statistical and mathematical foundations of deep learning have meant that qualitative research has almost necessarily had recourse to abstraction, an insistence on code as the object of analysis may advance and sustain humanistic engagement in that textual scholars can meaningfully contribute to the development of analytical frameworks that in turn inform actual research practices. 
‘Doing things’ with deep learning code, then, may very well mean helping to produce another possible future for language models, one that is more open and more available to both creative engagement and critical scrutiny.

\bibliography{bibtex/bib/ref.bib}{}
\bibliographystyle{IEEEtran}

\end{document}